\documentclass[11pt,a4paper]{article}
\usepackage[hyperref]{acl2021}
\usepackage{times} 
\usepackage{latexsym}

\usepackage{microtype}

\aclfinalcopy 

\usepackage{expex}
\usepackage{graphicx}

\lingset{
    textoffset=0.4em, 
    labeloffset=0.5em,
    exskip=0.5ex plus .1 ex minus .1 ex,
    interpartskip = 0.3ex plus .1 ex minus .1 ex}

\def\token#1{\texttt{\textsc{#1}}}
\def\DECL{\token{decl}}
\def\QUEST{\token{quest}}

\title{Transformers Generalize Linearly}

\author{Jackson Petty \\
  Yale University \\
  Department of Linguistics \\
  \texttt{jackson.petty@yale.edu} \\\And
  Robert Frank \\
  Yale University \\
  Department of Linguistics \\
  \texttt{bob.frank@yale.edu} \\}

\date{\today}

\begin{document}
\maketitle
\begin{abstract}
Natural language  exhibits patterns of hierarchically governed dependencies, in which relations between words are sensitive to syntactic structure rather than linear ordering. While recurrent network models often fail to generalize in a hierarchically sensitive way~\cite{mccoytrees} when trained on ambiguous data, the improvement in performance of newer Transformer language models~\cite{vaswani2017attention} on a range of syntactic benchmarks trained on large data sets \cite{goldberg2019assessing,warstadt2019investigating} opens the question of whether these models might exhibit hierarchical generalization in the face of impoverished data. In this paper we examine patterns of structural generalization for Transformer  sequence-to-sequence models and find that not only do Transformers fail to generalize  hierarchically across a wide variety of grammatical mapping tasks, but they exhibit an even stronger preference for linear generalization than comparable recurrent networks.
\end{abstract}

\section{Introduction}

One of the fundamental properties of human languages is their sensitivity to relations among elements that are not easily characterized in linear terms. In phenomena like subject-verb agreement or reflexive anaphora, the relationship between the agreeing verb and its agreement target or the reflexive pronoun and its antecedent is not governed by linear properties like adjacency or recency, but instead by the hierarchical organization of the sentence.
Similarly, the relationship between related sentences, which are represented in some grammatical theories as transformational operations or as lexical rules in others, is also governed by  hierarchical organization. English polar questions, for instance, involve the fronting of an auxiliary verb in the corresponding declarative to a sentence-initial position. %
Questions with complex subjects like~(\nextx a) demonstrate that the verb that is fronted in such cases is the determined by hierarchical prominence (i.e., \textsc{move-main} yielding (\nextx b)) and not linear considerations (\textsc{move-first} yielding (\nextx c) or \textsc{move-last} yielding (\nextx d)).
\pex<ex:question>
    \a {[The president who \textbf{can} smile] \textbf{will} lead [those who \textbf{would} sing].}
    \a \textbf{Will} the president who \textbf{can} smile \underline{\hspace{1em}} lead those who \textbf{would} sing?
    \a * \textbf{Can} the president who \underline{\hspace{1em}} smile \textbf{will} lead those who \textbf{would} sing?
    \a * \textbf{Would} the president who \textbf{can} smile \textbf{will} lead those who \underline{\hspace{1em}} sing?
    
\xe
\citet{chomsky1971problems} argues that, in spite of receiving little input of the form in~(\lastx b), which would unambiguously demonstrate the necessity for a hierarchically governed dependency, children uniformly generalize the process of question formation in a hierarchical fashion. Such consistent behavior suggests that humans possess an inherent bias of some sort towards hierarchical generalization (though see
\citet{ambridge2008structure} and \citet{perfors2011learnability} for
arguments against this view).
Replicating such a bias in
generalization would indicate the ability to mimic patterns
of human cognition and learning. 

Previous investigations of recurrent neural architectures have yielded some evidence for hierarchically-governed linguistic knowledge \cite{gulordava2018colorless, Marvin-2018-targeted,hu-etal-2020-systematic}.  Even greater success has been achieved with neural networks the incorporate explicit representation of syntactic structure \cite{kuncoro-etal-2018-lstms}. Architecturally-constrained models when trained without explicit information about syntactic structure show only modest benefits \cite{shen2018ordered,kim-etal-2019-unsupervised,merrill-etal-2019-finding}. However, all of these studies involve models that are trained on large quantities of text which may not be impoverished in domains that these benchmarks assess. As a result, it is unclear whether any apparent hierarchical behavior reported in these works is the effect of a bias for hierarchical generalization or the accumulation of patterns explicitly guided by the training data. \citet{mccoytrees} take a different tack: the training data  is carefully controlled so that hierarchical behavior can emerge  only if a model itself is biased to extract hierarchical  generalizations. Their experiments demonstrate that recurrent neural network seq2seq models show a clear preference for linear generalization.

The recently developed Transformer architecture has led to revolutionary advances across many areas of natural language processing, including
machine translation and question answering~\cite{vaswani2017attention, devlin2019bert}. Transformer-based models have also shown considerable success on benchmarks that appear to require the representation of hierarchical abstractions \cite{rogers2021primer,goldberg2019assessing, warstadt2019investigating}. Further, investigations of Transformers' representations of sentences \cite{hewitt-manning-2019-structural, lin-etal-2019-open} point to encodings of hierarchical syntactic structure. Yet, for the reasons noted above, it is difficult to conclude much about the inductive bias in the Transformer: they are trained on vast datasets, leaving open the question of the impact of inductive bias as opposed to training data (\citet{warstadt2020can}, but see \citet{van2019quantity}  for arguments that even massive data may not be sufficient). This paper contributes to our understanding by examining the degree to which the Transformer architecture is biased toward hierarchical generalization when the data underdetermine such generalization. Specifically, we study whether Transformers learning sequence-to-sequence mappings generalize in a structure sensitive way, and compare their performance with recurrent models.



\section{Experiments}

Our experiments involve a variety of
English-language transduction tasks that highlight 
hierarchically-governed patterns. For each task, the training data is ambiguous between a linear and hierarchical generalization. This allows us to evaluate performance on both a 
\textsc{test} set, drawn from
the same distribution as the training set, and a
\textsc{gen} set of data, that contains out-of-distribution data consistent only with hierarchical patterns of generalization.

We compare transformer models with a number of recurrent architectures (LSTMs and GRUs with no attention, with
additive attention \cite{bahdanau2016neural}, and with 
multiplicative attention \cite{luong2015effective}).
Transformer models follow their usual implementation with
self- and multi-headed attention. For each model type,
we perform 10 runs, initialized with different random initial seeds, and report median accuracy metrics. Recurrent
units are single-layer models, with hidden and embedding
dimensions of 256. Transformers are 4-headed, 3-layer models 
with hidden and embedding dimensions of 128. All models are
trained at a learning rate of 0.01 using SGD optimization for
100 epochs with early stopping.

\subsection{Polar Question Formation}

Our first task involves the process of question formation discussed earlier.
We borrow the formulation of this task from \citet{mccoytrees}: the training dataset consists of an input sentence (a simple declarative with relative clauses optionally modifying the subject and object), a transformation token,  \DECL\ or \QUEST, and an output sentence.  The transformation token  specifies what the form of the target output should be. Following the logic surrounding example (\getref{ex:question}), examples with subject-modifying relative clauses are never paired in the training data with the \QUEST\ transformation token. 
As a result, the network is not trained on sentences in which an auxiliary verb must be fronted past an intervening relative clause, and  the target generalization
is therefore  ambiguous between something akin to \textsc{move-main}
and \textsc{move-first}. While a network that acquires the \textsc{move-first} generalization will succeed on the
in-distribution \textsc{test} set consisting of  examples of the same structure as in the training data, it will fail on the \textsc{gen} set consisting of input sentences with 
subject-relative clauses and the \QUEST\ transformation.

\begin{figure}[t]
\includegraphics[width=\columnwidth]{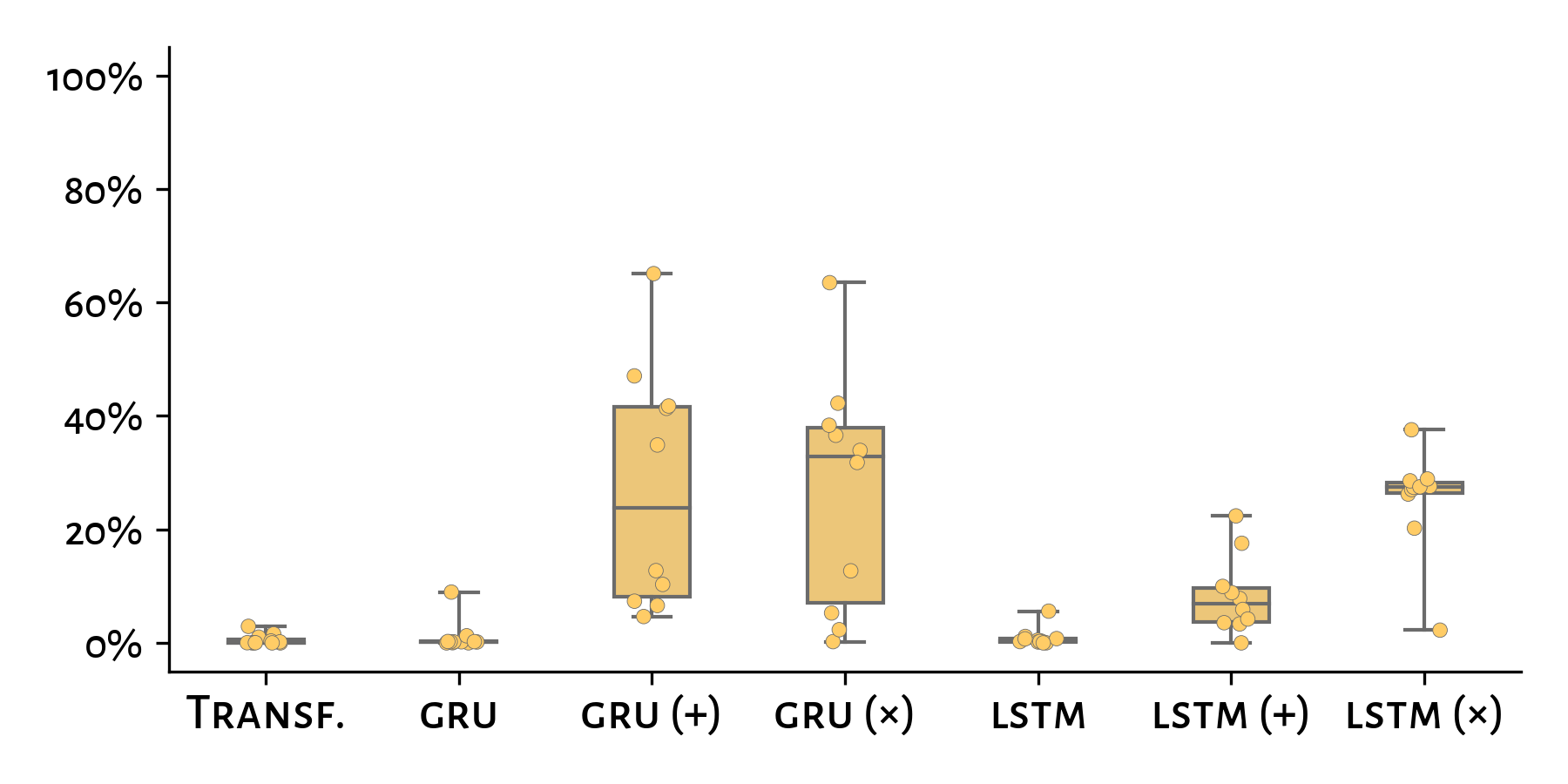}
\caption{Proportion of first-word predictions consistent with hierarchical generalization in the question \textsc{gen} set. A ($+$) denotes additive
attention, ($\times$), multiplicative. Horizontal bars denote max, median,
and min values.}
\label{fig:question}
\end{figure}

All trained network types performed well on the in-distribution
\textsc{test} set, attaining mean
full-sentence accuracies of at least 95\%. In contrast,
none of the models succeeded on the \textsc{gen} set in full sentence accuracy. Following \citet{mccoytrees}, we instead assess \textsc{gen} set performance using the more lenient metric of first-word accuracy. Since the \textsc{gen} set includes only sentences with distinct auxiliary verbs in the main and relative clauses, the identity of the first output word reveals whether the network has acquired a linear (\textsc{move-first}) or hierarchical (\textsc{move-main}) generalization.  Results are shown in 
Figure~\ref{fig:question}.
As noted in \citet{mccoytrees}, there is variation in performance among the different types of recurrent networks: GRUs with multiplicative attention 
achieved median accuracy of 32.9\%. Transformers exhibit the worst median performance among all
architectures surveyed, with a median first-word accuracy of
just 0.03\% and virtually no variability across different random initializations. Instead, Transformer
models overwhelmingly predicted sequences consistent with
a linear \textsc{move-first} rule on the \textsc{gen} set.
These results are robust across changes in learning
rate.

\subsection{Tense Reinflection}

Our second mapping task, again borrowed from \citet{mccoytrees} involves the reinflection of a sentence with a past tense verb into one with either a past or present tense verb.  Significantly, the English present tense involves structurally-conditioned agreement with the verb's subject. In complex expressions 
like~(\nextx a), distractor nouns with different number within the subject linearly separate the verb from the subject, but the grammatical agreement is nonetheless governed by a hierarchical \textsc{agree-subject} relation (predicting~(\nextx b)).
as opposed to an \textsc{agree-recent} relation 
(predicting~(\nextx c)).
\pex
    \a My newt near the elephants ran.
    \a My \textbf{newt} near the elephants \textbf{runs}.
    \a* My newt near the \textbf{elephants run}.

\begin{figure}[t]
\includegraphics[width=\columnwidth]{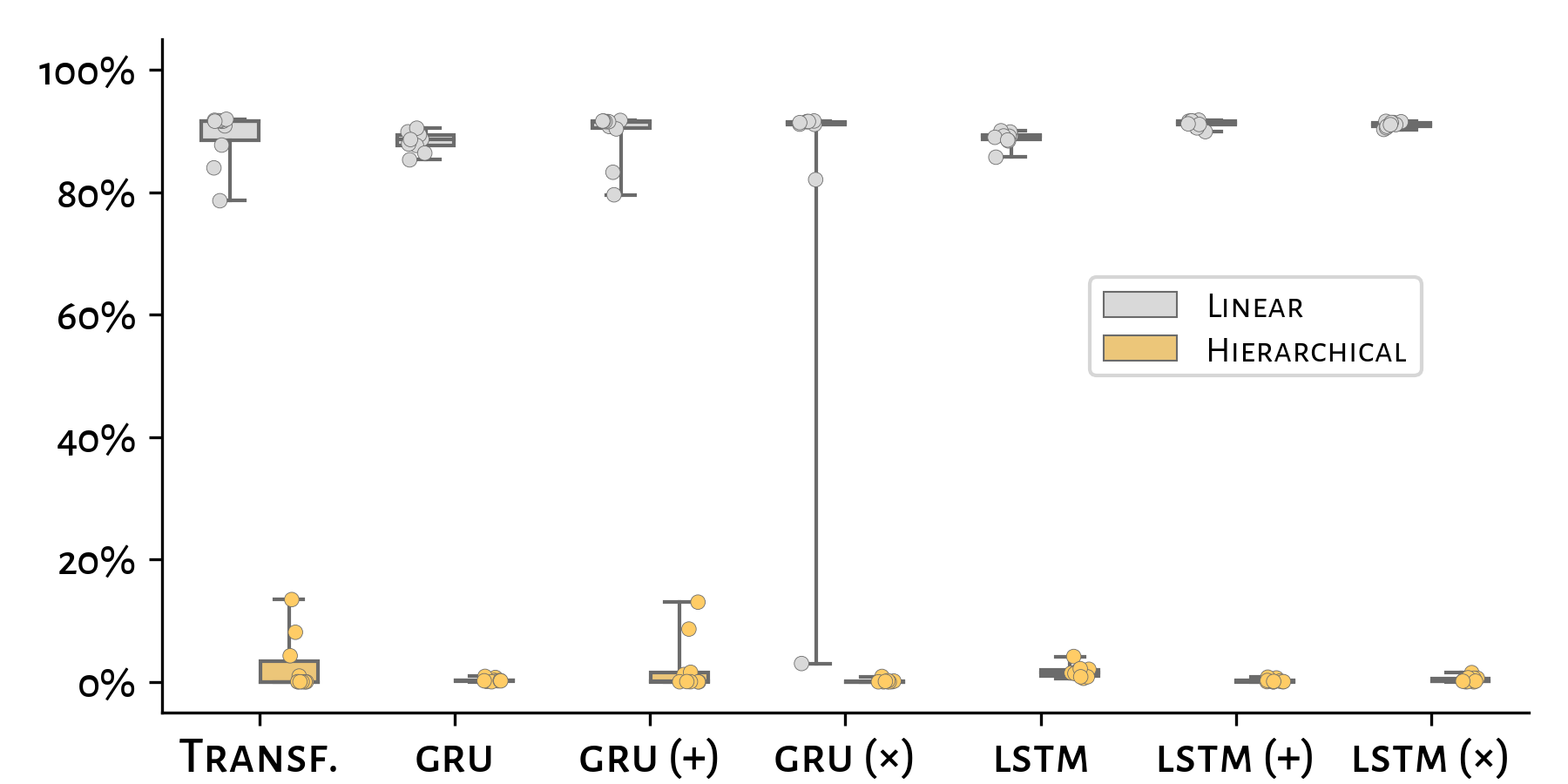}
\caption{Proportion of linear and hierarchical predictions on the reinflection \textsc{gen} set.} 
\label{fig:tense}
\end{figure}

\xe
Our datasets consist of past-tense English
sentences as inputs, optionally with prepositional phrases or relative clauses modifying the
subject or object, along with \textsc{pres} and \textsc{past} transformation tokens that indicate the form of the target output. For training and in-distribution test data, examples with the \textsc{pres} token do not have modified subjects, so that the reinflection mapping is
ambiguous between \textsc{agree-subject} and
\textsc{agree-recent}. In contrast, the \textsc{gen} set includes sentences where the two rules make different
predictions (modified subjects with distractor having distinct number).  Results are shown in Figure~\ref{fig:tense}. Like the recurrent architectures, Transformers systematically fail to exhibit hierarchical in favor of linear generalization.

\subsection{Negation}

Our third task involves the conversion of an affirmative sentence into a negative one.  Negation requires the insertion of the negative marker ``not'' immediately
prior to the main verb.
\pex
    \a The bird will sing.
    \a The bird will \textbf{not} sing.
\xe
When an adverbial clause is placed before or after the
main clause~(\nextx), the main verb is no longer consistently the linearly first or last verb in the sentence. 
\pex
    \a The bird will sing because the cat will swim.
     \a The bird will \textbf{not} sing because the cat will swim.
    \a Because the cat will swim the bird will \textbf{not} sing.
\xe
Our dataset consists of affirmative sentences, with adverbial clauses optionally preceding or following the main clause. These are transformed either into (identical) affirmatives or corresponding negatives. The training and in-distribution test set excludes sentences with initial adverbial clauses that must be mapped to negatives. 
As a result, this data set is ambiguous between a linear
\textsc{neg-first} generalization and a hierarchical \textsc{neg-main}.
This ambiguity is resolved in the \textsc{gen} set, which
contains sentences with preceding adverbials that must be converted into negative sentences, following the \textsc{neg-main} generalization.

All models, including the Transformer, perform exceedingly well on in-distribution data,
attaining near-ceiling full-sentence accuracy on the
\textsc{test} set. By contrast, all models, again including the Transformer, fail uniformly
on the \textsc{gen} set, attaining near-zero performance even using a
more forgiving metric looking only at
correct placement of the negative marker.
Closer examination of the model outputs on the \textsc{gen}
set reveals that networks of all sorts overwhelmingly produce predictions consistent with the linear generalization (\textsc{neg-first}).


\subsection{Reflexive Anaphoric Interpretation}

Our final task, similar to that of~\citet{kim-linzen-2020-cogs} and \citet{frank-petty-2020-sequence}, involves the semantic parsing of a sequence into a predicate calculus representation, as in~(\nextx).
\ex
    Alice sees Bob $\to$ \textsc{see}(\textsc{alice}, \textsc{bob})
\xe
For entities whose meaning is context-independent, like
nouns or verbs, this task involves learning a combination
of token correspondence and form composition. As \citet{frank-petty-2020-sequence} note, reflexive
anaphora like ``herself'' present a challenge since their
meaning is not context-independent but rather conditioned
on a linguistically-determined antecedent. In sentences with complex subjects, like that in~(\nextx) with
a prepositional phrase modifier, the identification
of the correct antecedent for the anaphor is conditioned
not by the linear distance between a potential antecedent and the reflexive but rather by the hierarchical relation between the antecedent and reflexive.
\ex  The boy by the king sees himself $\to$ \\
    \textsc{see}(\textsc{boy}, \textsc{boy}) $\land$
    \textsc{by}(\textsc{boy}, \textsc{king})
\xe

Our in-distribution data consists of sentences, transitive and
intransitive, paired with predicate calculus representations
of their meanings. Input sentences in this set may have complex subjects or the reflexive objects (``himself'' or ``herself''), but not both. As a result, the training and \textsc{test} data does not disambiguate whether the reflexive is co-referent with the grammatical subject or the noun phrase immediately preceding the verb.
The \textsc{gen} set contains only sentences reflexive objects and complex subjects containing prepositional phrases, and therefore serves to distinguish between the linear and hierarchical generalizations. 

\begin{figure}[t]
\includegraphics[width=\columnwidth]{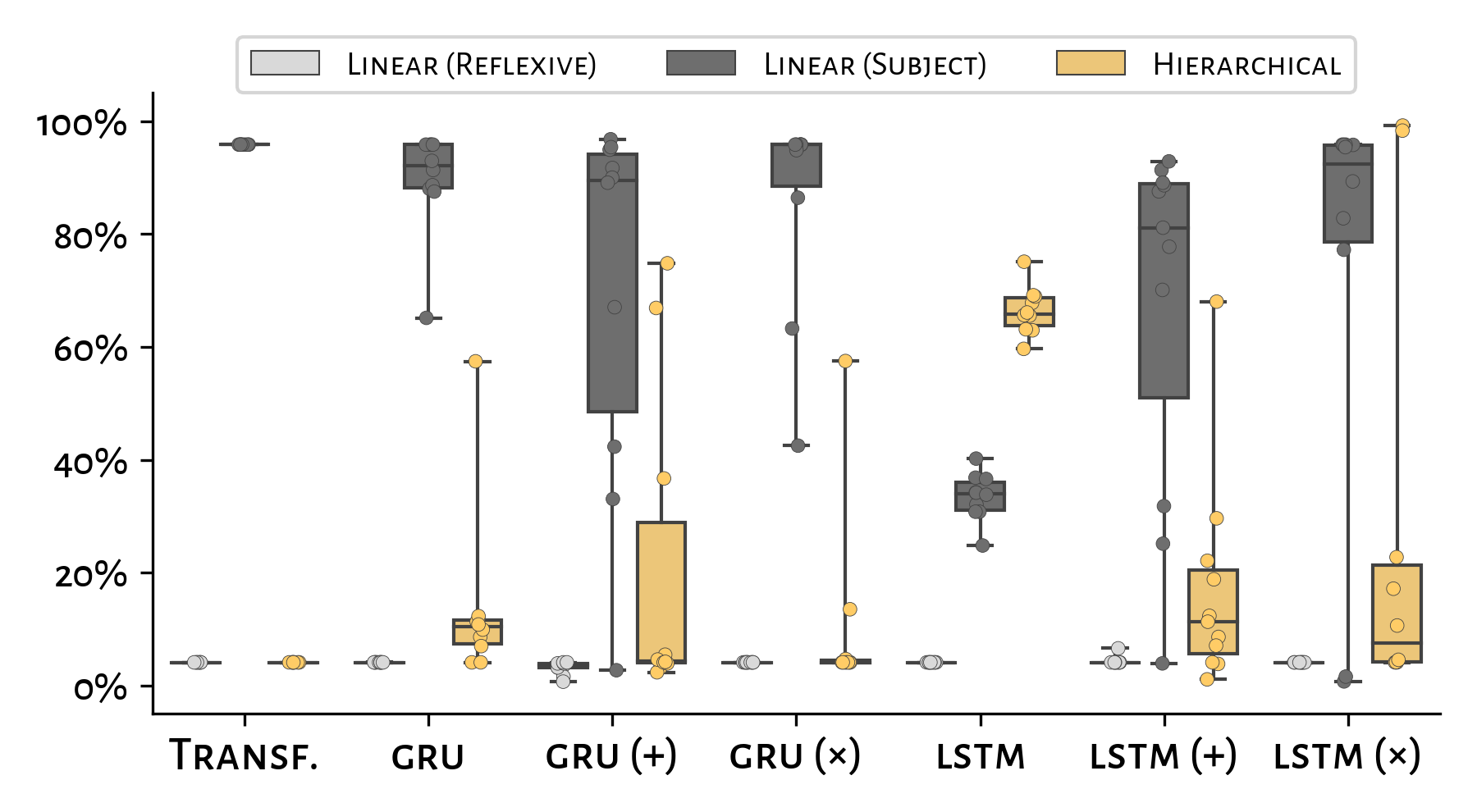}
\vspace*{-4ex}
\caption{Proportion of reflexive-linear, subject-linear, and hierarchical predictions in the anaphora \textsc{gen} set.}
\label{fig:anaphora}
\end{figure}

All models examined perform well on the \textsc{test} set,
attaining median full sequence accuracy of 100\%.
Results on the \textsc{gen} set, as shown in 
Figure~\ref{fig:anaphora}, are more varied.
We categorize the predictions made by the network into three
distinct classes: subject-verb linear, where the model
interprets the subject of the verb as being the linearly most
recent noun (incompatible with the training data); reflexive linear, where the model interprets
the antecedent of the reflexive as being the linearly most recent
noun (compatible with the training set); and hierarchical, where the model correctly interprets
both the subject and antecedent in a manner consistent with
the hierarchical structure of the sentence (also compatible with training). Transformers and GRU models overwhelming make predictions
consistent with reflexive linearity.
LSTMs are more varied, with inattentive LSTMs attaining
the highest hierarchical scores of all network types with
a median performance of 65.8\%. 

\section{Conclusion}

Transformers have shown great success on syntactic benchmarks. Is this because the architecture has useful syntactic biases, or is it because cues to hierarchical structure are present in their training data? 
Our results find no evidence for the former, suggesting that their syntactic successes can mainly be attributed to their ability to leverage massive training sets rather than linguistically-relevant architectural biases. 
Though the Transformer models studied here were the best performers on
in-distribution data across all tasks, their strong preference for  linear over 
hierarchical generalization suggests an explanation for their poor performance on 
tasks requiring structural generalization~\citep{kim-linzen-2020-cogs} 
despite their promise in other syntactically sensitive tasks. 
Finally, we note that the preference we have observed for linear generalization is consistent with
previous theoretical work on the (limited) expressive power of Transformers
\cite{hahn-transformer,merrill2019thesis}.

\bibliographystyle{acl_natbib}
\bibliography{anthology,acl2021,sources}


\end{document}